\documentclass[letterpaper, 10 pt, journal]{IEEEtran}
\IEEEoverridecommandlockouts                              
\usepackage[english]{babel}
\usepackage[inkscapelatex=false]{svg}

\addto\captionsenglish{\renewcommand{\figurename}{Fig.}}

\usepackage{cite}
\usepackage{amsmath,amssymb,amsfonts}
\usepackage{algorithmic}
\usepackage{graphicx}
\usepackage{textcomp}
\usepackage{xcolor}
\usepackage{array}
\usepackage{mathtools}
\def\BibTeX{{\rm B\kern-.05em{\sc i\kern-.025em b}\kern-.08em
    T\kern-.1667em\lower.7ex\hbox{E}\kern-.125emX}}

\newcolumntype{P}[1]{>{\centering\arraybackslash}m{#1}}

\begin{document}

\title{Electrostatic Clutch-Based Mechanical Multiplexer \\with Increased Force Capacity}





\author{Timothy E. Amish$^{1}$, Jeffrey T. Auletta$^{2}$, Chad C. Kessens$^{2}$, Joshua R. Smith$^{1,3}$, and Jeffrey I. Lipton$^{4*}$
\thanks{$^{1}$ Dept of Electrical and Computer Engineering, University of Washington, Seattle, WA, 98195 USA }%
\thanks{$^{2}$ US Army Research Directorate, DEVCOM Army Research Laboratory, Aberdeen Proving Ground, MD 21005 USA }%
\thanks{$^{3}$ Paul G Allen School of Computer Science and Engineering, University of Washington, Seattle, WA, 98195 USA }%
\thanks{$^{4}$ Mechanical and Industrial Engineering Department of Northeastern University, Boston, MA, 02115 USA }%
\thanks{$^{*}$ {\tt\small j.lipton@northeastern.edu}}%
\vspace{-.35cm}
}

\maketitle

\begin{abstract}


As robotic systems become increasingly articulated, conventional actuation still dedicates one motor to each degree of freedom (DoF). Mechanical multiplexers address this limitation by allowing a single motor to control multiple outputs through clutches, reducing the number of required motors. However, previous multiplexers have relied on bulky mechanical clutch designs, limiting their development. This study presents an electrostatic capstan clutch-based transmission architecture that enables high-force mechanical multiplexing with independent, simultaneous, and fully actuated control of multiple outputs from a single motor. Our transmission implements four fully-actuated linear outputs, achieving individual output forces of up to 212 N and output speeds of up to 69.5~mm/s. We demonstrate our transmission on a commercial tendon-driven hand, where sequentially allocating system-wide torque capacity to individual outputs increased vertical grip strength by 4.09$\times$ and raised horizontal carrying capacity to 111.2 N, the highest reported among five-fingered tendon-driven robotic hands. These results demonstrate that electrostatic clutch-based mechanical multiplexing enables high-force, independent, simultaneous, and fully actuated control while overcoming the limitations of previous mechanical multiplexers.

\end{abstract}

\vspace{-2mm}
\section{Introduction}

Highly articulated robotic systems capable of dexterous manipulation often require both a large number of degrees of freedom (DoF) and degrees of actuation (DoA). The current paradigm for robotic actuation is to dedicate a motor to each joint or tendon. Dedicating one motor per joint scales poorly as robots become more articulated, increasing system cost, weight, and power consumption, while often requiring actuators to be distributed throughout the robotic structure. Relying on a large number of actuators for robot actuation has been increasingly criticized by some designers and researchers \cite{Plooij-2015-lock,Billard-2019-trends,Transeth-2009-snake,Sun-2022-DesignPrinciples}

Underactuated systems address this problem by reducing actuator count, but do so at the expense of reduced DoF control. Mechanical multiplexing transmissions use clutches to enable a single actuator to operate multiple outputs, preserving control over multiple DoF. A primary challenge in the implementation of mechanical multiplexing has been the clutch technology itself. Conventional mechanical clutches are often bulky and slow, limiting scalability and performance. 

This paper presents a novel mechanical multiplexing transmission architecture that enables a single actuator both independent and simultaneous actuation of multiple outputs while maintaining fully-actuated control of each DoF through dynamic power allocation using previously developed electrostatic rotary capstan clutches \cite{Amish-2024-JRCC}. 

\begin{figure}[t]
\centering
\includegraphics[width=0.99\linewidth]{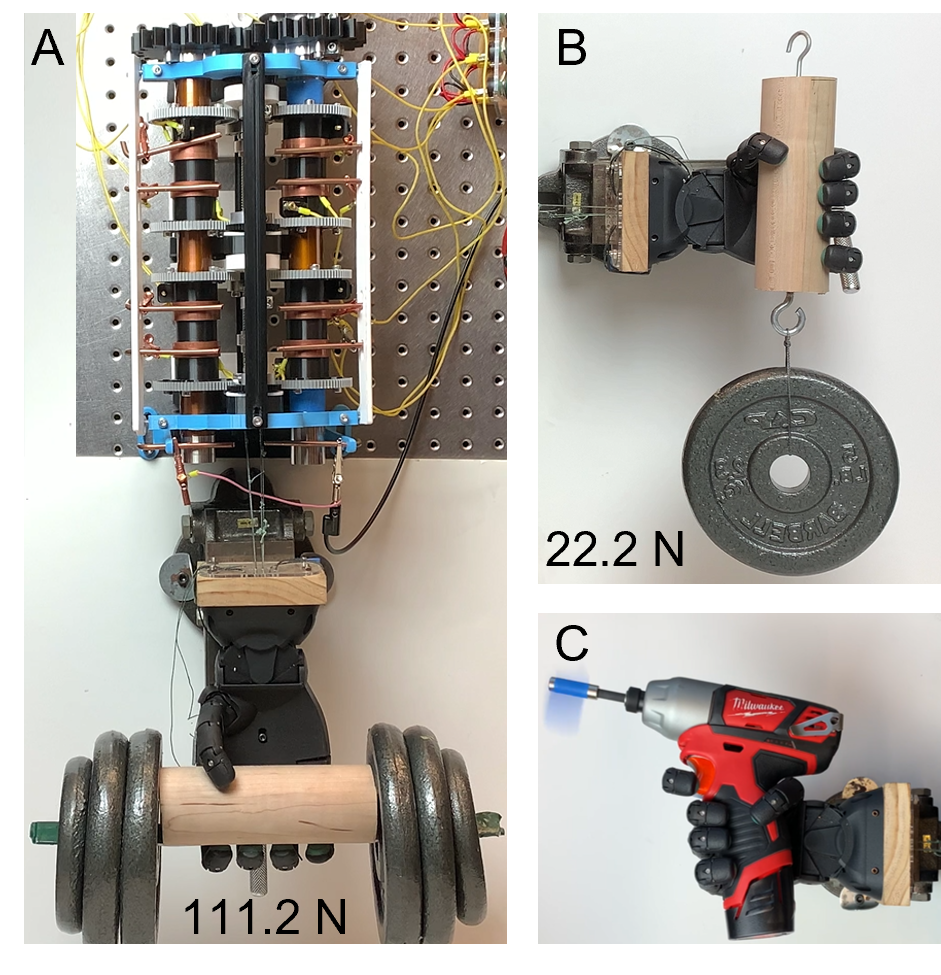}
\vspace{-8mm}
\caption{\label{fig:Hero} In this work, we show how our novel multiplexing architecture enables a single motor to: A) Increase horizontal holding capacity to 111.2~N, the highest recorded value compared to other five-fingered tendon-driven robotic hands. B) Increase vertical grip strength over conventional multi-actuator architectures by up to a factor of 4.09 to 22.2~N with SISO multiplexing. C) Demonstrate the functionality of our system by using SIMO multiplexing to grasp a drill, SISO multiplexing to increase vertical grip strength from 55.1~N to 118.76~N, and to actuate a modified trigger.}
\vspace{-5mm}
\end{figure}

\vspace{2mm}
Contributions Summary: 

\begin{itemize}
    \item Present an electrostatic capstan clutch-based mechanical multiplexer that enables independent and simultaneous actuation of multiple outputs while preserving fully-actuated control of each DoF and requiring no mechanical reconfiguration.
    \item Show how sequential output switching with mechanical multiplexing can increase individual output forces and system-wide force output.
    \item Use our mechanical multiplexer to increase grip strength of a robotic hand, demonstrating the highest horizontal holding capacity across five-fingered tendon-driven robotic hands.
    \item Provide system-level models for different mechanical multiplexer operating modes. 
\end{itemize}

Experimental validation demonstrates independent, simultaneous, and fully-actuated DoF control of four outputs using a single actuator, achieving individual output forces of up to 212 N, 152~N dynamic response time as low as 22.9~ms, and output speeds of up to 69.5 mm/s. A system-level model is developed to identify operating regimes and provide predictive design capability, with force-capacity predictions validated experimentally. Dynamic performance is also experimentally characterized. The proposed architecture is demonstrated using a commercial robotic hand, where four independent outputs are controlled using a single motor, as shown in Fig.~\ref{fig:Hero}. We further demonstrate how mechanical multiplexing can increase grasp strength by allowing each finger to sequentially apply the full shared actuator force before holding position, effectively increasing system-wide force over time. Although demonstrated using a robotic hand, the proposed architecture is applicable to a broad range of highly articulated robotic systems. Potential applications include robotic manipulators, actuator arrays, haptic displays \cite{Zhang-2018-Display}, and other high-DoF platforms.

\section{Mechanical Multiplexing and Fully-Actuated DoF Control}
\subsection{Mechanical Multiplexing}
A conventional fully-actuated robotic system assigns a dedicated actuator to each degree of freedom (DoF) \cite{He-2019-underactuated,Zuo-2020-Mplex-SurgicalRobot,Kim-2022-Switchable-Cable,Sun-2022-DesignPrinciples}. Although this architecture provides independent control, it scales poorly as systems become increasingly articulated, as motors are typically the most expensive, power-consuming, and heavy components in a robot \cite{Plooij-2015-lock,Billard-2019-trends,Sun-2022-DesignPrinciples}. Underactuated systems reduce actuator count but sacrifice independent control of some DoF \cite{He-2019-underactuated,Patrick-2024-Underactuated}.

Mechanical multiplexing seeks to reduce actuator count while preserving independent control by enabling a single actuator to service multiple output channels through clutching mechanisms. To establish terminology, let $N$ denote the number of outputs and $A$ the number of actuators. Conventional architectures satisfy $N\leq A$, whereas mechanically multiplexed systems satisfy $N>A$.

Mechanical multiplexers can be classified by how actuator resources are distributed among outputs. In this work, we define two types of operating modes: single-input-single-output (SISO) multiplexing and single-input-multiple-output (SIMO) multiplexing.

\begin{figure}[t]
\centering
\includegraphics[width=0.99\linewidth]{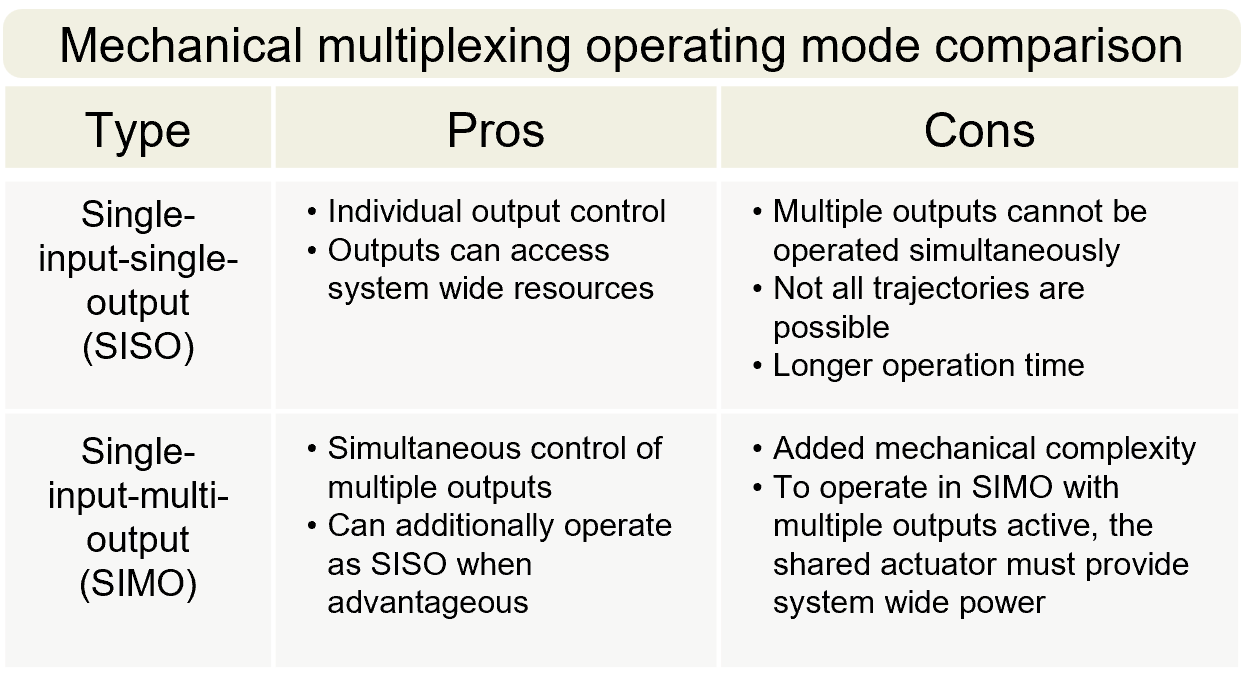}
\vspace{-8mm}
\caption{\label{fig:FunctionalityComparison} Chart comparing SISO and SIMO mechanical multiplexing.}
\vspace{-5mm}
\end{figure}

\subsection{SISO and SIMO Multiplexing}

In SISO multiplexing, a single actuator sequentially services multiple outputs by connecting to only one output channel at a time. This allows the full actuator capacity to be allocated to an individual output, increasing peak output force. However, because outputs are serviced sequentially, operation can be slower and path trajectories requiring simultaneous multi-DoF motion are not possible \cite{Patrick-2024-Underactuated,He-2019-underactuated}.

In SIMO multiplexing, a single actuator independently and simultaneously controls multiple outputs. SIMO enables coordinated multi-DoF motion but requires the actuator capacity to be shared among active outputs. Consequently, SIMO generally provides greater operational functionality. It follows that if a system accomplishes SIMO multiplexing it has the function of SISO multiplexing as only one output could be engaged. A qualitative comparison of these operating modes is shown in Fig.~\ref{fig:FunctionalityComparison}.

\subsection{Prior Mechanical Multiplexers}

Mechanical multiplexing systems have been demonstrated using a variety of clutch technologies \cite{Zhang-2018-Display,Strong-1970-electro-display,Lancaster-2022-effector,Aukes-2014-hand,Patrick-2024-Underactuated,fang-2021-earthworm,Zuo-2020-Mplex-SurgicalRobot,Kim-2022-Switchable-Cable,Xu-2024-MuxHand}. Most reported systems can only operate in SISO mode \cite{Zhang-2018-Display,Strong-1970-electro-display,Lancaster-2022-effector,Aukes-2014-hand,Patrick-2024-Underactuated,fang-2021-earthworm,Zuo-2020-Mplex-SurgicalRobot,Kim-2022-Switchable-Cable,Xu-2024-MuxHand}. The comparatively few SIMO-capable systems are generally more difficult to construct and have relied on manual control to mechanically reconfigure \cite{Mathieu-2013-MechanicallyProgrammedMultiplexHand,armatron} or employ underactuated architectures that do not preserve fully-actuated control of individual DoF \cite{Wei-2023-SoftRobotic-Multiplexing, Colgate-2026-Hand}. In addition, multiplexers with underactuated DoFs rely on additional mechanisms to operate, such as requiring a spring to reset to a starting configuration \cite{Colgate-2026-Hand,Wei-2023-SoftRobotic-Multiplexing}.




The electrically controlled robot transmission architecture presented here only uses a single motor for SISO and SIMO operation while preserving fully-actuated control of each DoF and requiring no mechanical reconfiguration. A comparison of the functionality of representative multiplexing architectures is shown in Fig.~\ref{fig:FunctionalityComparison}.

\section{Single Mechanical Multiplexer Unit Full Actuation Design and Operation}


\begin{figure*}[t]
\centering
\includegraphics[width=0.8\linewidth]{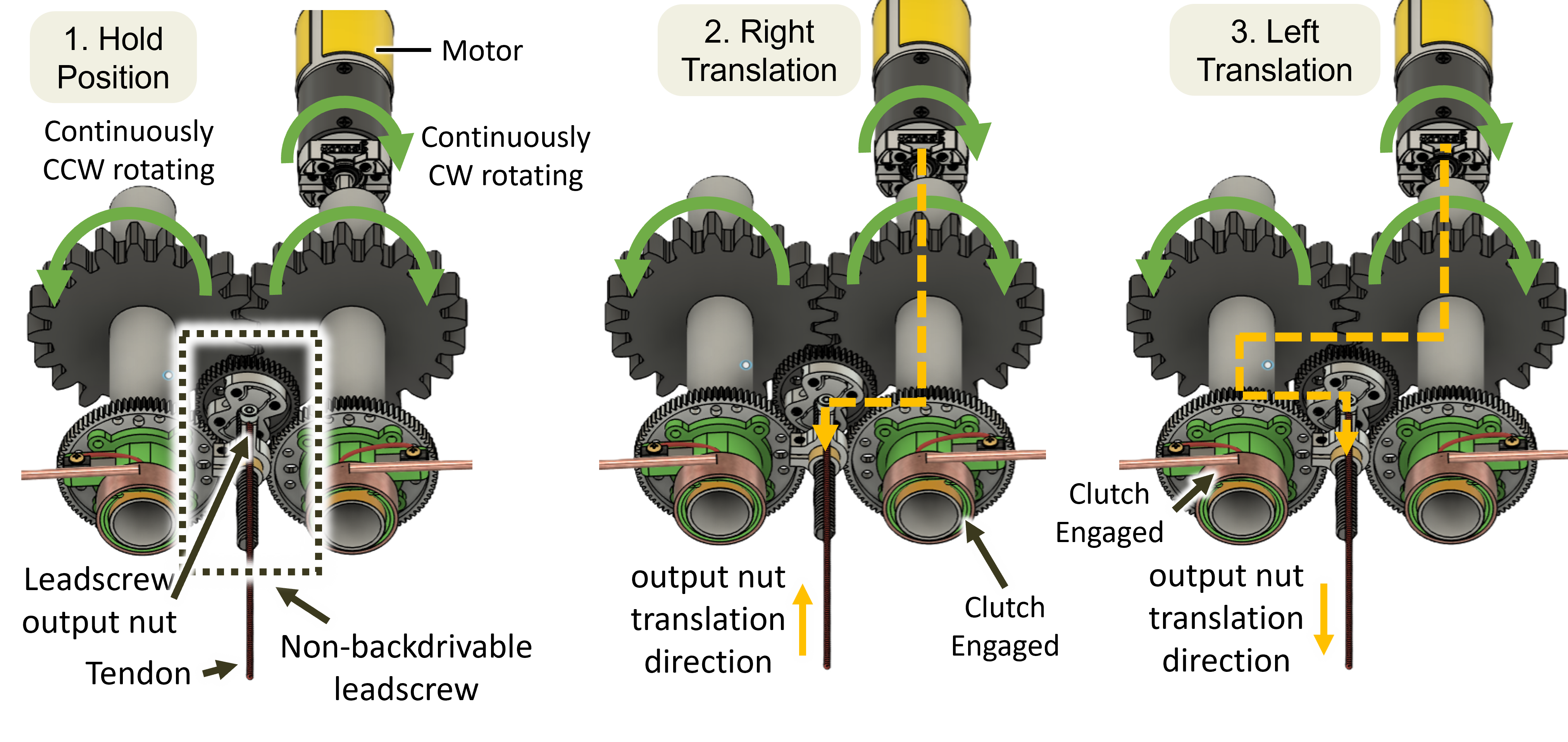}
\vspace{-3mm}
\caption{\label{fig:single-unit} Implementation of a single unit to control one DoF, with a leadscrew nut acting as the output and capable of rightward translation, leftward translation and holding position. A single motor continuously rotates one input shaft CW and counter rotates the second input shaft CCW through a gear.  When the right translation clutch is engaged, it couples the CW rotating shaft and leadscrew, causing a rightward translation. The left translation clutch couples the leadscrew to the CCW rotating clutch, resulting in leftward translation. The leadscrew is not backdrivable and maintains position when neither clutch is engaged.}
\end{figure*}

For fully-actuated DoF control, three functions are necessary. The first function is the ability to control the position. The last two functions are controlling both movement directions; right and left for translational movement and clockwise (CW) and counter-clockwise (CCW) for rotational movement \cite{Fully-actuated-definition}. Given the use of $n$ clutches that are engaged or disengaged, there is the possibility of $2^n$ states. It follows that if we want to fully actuate one DoF, at least two clutches are required. A single output is operated with two clutches and a leadscrew. 

A diagram of the operation and construction of our mechanical multiplexing architecture is shown in Fig.~\ref{fig:single-unit}. These single units will be added in parallel to form the complete multiplexer with four outputs. To provide mechanical power to the entire system, a single motor is continuously rotating one input shaft CW and is geared to a second input shaft in parallel to continuously rotate CCW. 

The first function of controlling position is accomplished with the non-backdrivability property of a leadscrew. Non-backdrivable mechanisms provide passive holding without actuator power and prevents position change during power loss \cite{Plooij-2015-lock}. Non-backdrivability has the drawback of obscuring force transparency but can be addressed through direct force sensing and feedback \cite{Hiroshi-2013-AddressingNonbackdriability}. When neither clutch is engaged, the non-backdrivability of the leadscrew causes it to ``self-lock'' and prevents further linear motion if the friction angle is greater than the lead angle \cite{bhandari-2007-leadscrew}. The two clutches control the remaining functions of both rightward and leftward motion. 

When switched ON, the rightward clutch begins to engage, coupling the CW input shaft and causing a rightward translation of the output leadscrew nut. Switching the leftward clutch ON, couples the CCW input shaft and leadscrew, causing a leftward translation of the output leadscrew nut. The mechanical power transfer for both cases is shown by the orange dotted line in Fig.~\ref{fig:single-unit}. In both cases, it is important that both clutches are not engaged at the same time. Engaging both clutches couple the CW and CCW input shafts and prevents the motor from rotating. When the rightward clutch is engaged, the leftward clutch is disengaged to freely rotate on its input shaft (with residual friction from the clutch casing) and vise versa.


One motor is able to control four outputs with fully-actuated DoF, by tiling four pairs of electrostatic clutches and four leadscrews, along the input shafts. For each additional DoA, two clutches and a leadscrew are tiled in parallel along the rotating input shafts. Tiling units in parallel allows for the independent and simultaneous control needed for SIMO multiplexing. 



\section{System-Level Modeling and Operating Regime}
\label{sec:SystemModel}

To establish predictive design capability and clarify practical applicability, we develop a system-level model that maps motor torque to output force, and defines operating regimes for SISO and SIMO multiplexing. This analysis exposes the physical limits of the architecture and distinguishes implementation dependent losses from fundamental transmission constraints.

\subsection{Output Force Transmission Model}

In our system, each output channel consists of a pair of clutches and a leadscrew. For each output channel $i$, an electrostatic clutch is modeled as an electrically controlled torque limiter that transmits torque up to a maximum holding torque $\tau_{c,i,\max}$ determined by the clutch voltage and wrap geometry as described in \cite{Amish-2024-JRCC,Colgate-2026-Hand}. 
A non-active channel contributes a small residual drag $\tau_{c,drag}$ as the disengaged clutch casing slides along the input shaft. The output force transmitted in an active channel $i$ is the transmitted clutch torque $\tau_{c,i}$ multiplied by a torque-to-force conversion factor $K_i$ that represents the dependence on the gear ratio, gear efficiency, leadscrew geometry, and leadscrew efficiency \cite{bhandari-2007-leadscrew}. 

\begin{equation}
F_{i} = K_i \cdot\tau_{c,i}.
\end{equation}


Let $\tau_{\Delta,actuators}$ denote the total summed torque available in the system. For our mechanical multiplexer this corresponds to our single motor. For a traditional system, it is the summed total of maximum torque for all the actuators used. Let $N$ denote the number of output channels, $N_a$ simultaneously active channels with the number of disengaged clutches $(2N-N_a)$ that contribute to drag. The remaining torque available $\tau_{\mathrm{available}}$ becomes

\begin{equation}
\tau_{\mathrm{available}} = 
\tau_{\Delta,actuators} - \sum_{k=1}^{N_a} \tau_{c,k}
- (2N-N_a)\tau_{c,drag}.
\label{eq:torq_available}
\end{equation}


\subsection{SIMO / SISO Operation}
in the common case where torque divides approximately evenly across $N_a$ active channels,

\begin{equation}
F_i \approx \frac{K_i}{N_a} \cdot [
\tau_{\Delta,motor}
- (2N-N_a)\tau_{c,drag}].
\end{equation}

Under SISO operation only one channel is active $N_a =1$ and compared to the output force when all channels are active with SIMO $N_a = N$, the \textit{per-channel} force ratio becomes,

\begin{equation}
\frac{F^{\mathrm{SISO}}_i}{F^{\mathrm{SIMO}}_i} \approx 
\frac{N(\tau_{\Delta,actuators} - (2N-1)\tau_{c,drag}}{\tau_{\Delta,actuators - N\tau_{c,drag}}}.
\label{eq:Fi_ratio}
\end{equation}

highlighting two important scaling behaviors: (i) peak force
decreases with increasing concurrency, and (ii) efficiency
losses grow with the number of disengaged clutches due to
accumulated drag torque.


\subsection{SISO Increased Force Capacity}

Using \ref{eq:Fi_ratio} and taking the drag torque as $\approx0$, the individual force ratio is reduced to the number of output channels $N$

\begin{equation}
\frac{F^{\mathrm{SISO}}_i}{F^{\mathrm{SIMO}}_i} \xrightarrow[]{\tau_{c,drag} \approx 0} N.
\label{eq:SISO/SIMO}
\end{equation}

In addition, in our case with the leadscrew, the output will be maintained, which means that over time the total output forces throughout the system can also be increased by $N$ in the ideal case. Using SISO to increase the system wide output force comes at the operational tradeoffs of 1) longer operation time 2) reduced system trajectories \cite{Patrick-2024-Underactuated}.

With the same assumption that drag torque is $\approx0~N$, the increase in force capacity is the same compared to conventional multi-actuator architectures with an identical drive train. Conventional multi-actuator architectures use a dedicated actuator $A$ for each output channel. The comparison of individual output forces for one channel with traditional architectures becomes the same as \eqref{eq:SISO/SIMO}, since $N\leq A$ for traditional architectures. 

\subsection{Operating Regime for SISO and SIMO}
This model provides explicit design relationships between actuator torque, clutch parameters, and achievable multi-channel force allocation, enabling system sizing and operating regime selection to determine when to use SISO vs SIMO. When the actuator is sized to match the total system demand, sequential routing enables a single channel to access nearly the full motor capability. In the ideal symmetric case, concentrating the power onto one of the $N$ outputs yields approximately a $N$-fold increase in the peak force per-channel relative to simultaneous actuation. Tasks characterized by many forces that do not exceed system capacity \eqref{eq:torq_available}, favor SIMO operation for coordinated motion, while tasks dominated by intermittent high-force requirements require SISO multiplexing. Section~\ref{sec:SIMO-Positioning-SISO-Increase} demonstrates a hybrid approach that employs SIMO for configuration and SISO to increase the output forces.

\section{Mechanical Multiplexer Implementation}
\subsection{Clutch Design and Construction}
\label{sec:ClutchDetails}
\begin{figure}[t]
\centering
\includegraphics[width=0.98\linewidth]{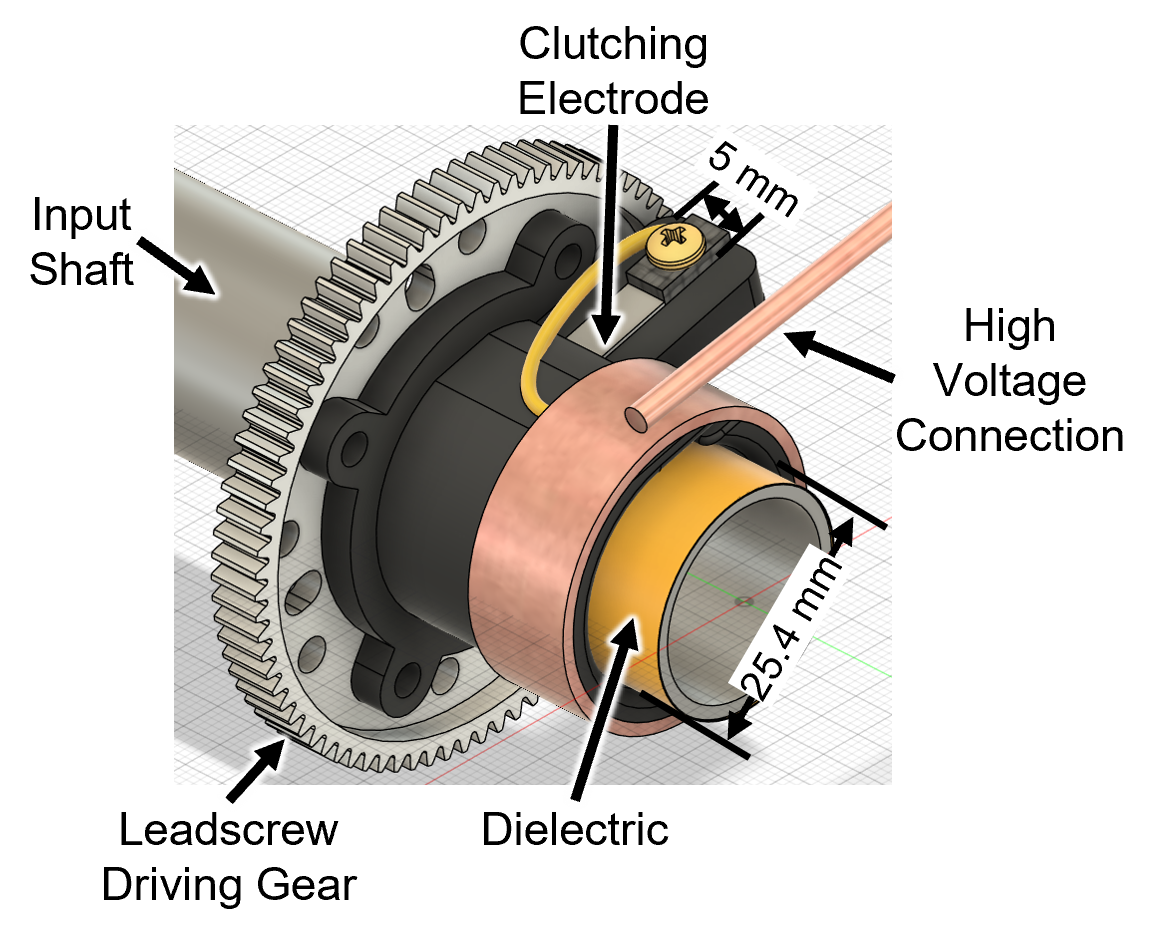}
\vspace{-3mm}
\caption{\label{fig:Clutch-Cad} Implementation of a electrostatic capstan clutch described in \cite{Amish-2024-JRCC}. When a voltage is applied between the clutch electrode and input shaft, the clutch electrode electrostatically adheres to the dielectric surface coating the input shaft, causing the clutch to rotate along with the input shaft.}
\end{figure}

Electrostatic clutches produce an attractive force when a voltage is applied across two conductive surfaces separated by a dielectric \cite{Guo-2020-tech-review}, mechanically coupling the surfaces and allowing transmission of mechanical force through friction. Electrostatic clutches were selected for this work due to their fast response times, precise electrical control, and high holding forces \cite{Patrick-2024-Underactuated,Diller-2018-DesignParam,Hinchet-2020-High-force,Choi-2024-HighPreformanceHighBandwidth,Wei-2023-SoftRobotic-Multiplexing,Krimsky-2024-Recycling,Bekir-2025-Multilayer}, although they are typically designed for coupling linear motion. We previously developed an electrostatic rotary capstan clutch (JRCC) capable of generating high holding torques using the capstan and Johnsen-Rahbek effects \cite{Amish-2024-JRCC}. It was shown to be capable of significantly higher holding torques compared to other rotary designs \cite{Johnsen-1923,Wei-2023-SoftRobotic-Multiplexing,Feizi-2023-SmartRotaryESclutch,Kornbaum-2024-RotaryESClutch,Diller-2022-Patent1,Diller-2025-Patent2}. The theory and operation of electrostatic capstan clutches is expanded on in \cite{Colgate-2026-Hand,Bekir-2025-Multilayer}. In this paper, we implement the JRCC design \cite{Amish-2024-JRCC} as shown in Fig.~\ref{fig:Clutch-Cad}. Each clutch is designed for a 25.4~mm diameter stainless steel input shaft (acting as one electrode), with an outside layer of 55~$\mu m$ polybenzimidazole film (PBI, $\varepsilon_d$ = 3.9) adhered using double-sided conductive carbon tape. The second electrode is formed from a 5~mm wide stainless steel (SS) shim, 0.0127~mm thick, wrapped around the dielectric coated shaft 4.72~radians. To reinforce the attachment point, a 5~mm x 10~mm carbon fiber sheet is epoxied onto the stainless steel band and a 3~mm bolt through the carbon fiber sheet and the band fixes the electrode to the 3D-printed clutch casing. PBI dielectric breakdown occurs at 1200~V, so the clutch ON voltage was set to 1000~V.

\subsection{Mechanical Multiplexer Construction}
\label{sec:MechanicalMultiplexerConstruction}

Our implementation of our mechanical multiplexing architecture detailed above is given here. 

When the input torque is limited to a specific value, a Yellowjacket 5202-2402-0100 motor is attached to one input shaft and set to continuously rotate CW and rotates a second parallel input shaft CCW using a gear. When an input torque is not specified, an AK60-6 motor is used for its higher rpm. 

Both input shafts are hollow stainless steel with a diameter of 25.4~mm and a length of 254~mm. A single DoA is implemented using a leadscrew, with the leadscrew nut acting as the output, and two electrostatic capstan clutches (shown in Fig.~\ref{fig:Clutch-Cad}) to control the rotation direction of the leadscrew. The leadscrew (8~mm lead, 4 start, 2~mm pitch) is rotated using a 46 tooth, 36.8~mm pitch diameter gear meshed with 80 tooth, 65.6~mm pitch diameter gears bolted to the clutches. All components are fixed in place to an optics table using PLA 3D printed brackets. No rolling-element bearings were incorporated in the transmission. The objective of this work was to demonstrate multiplexing functionality rather than optimize efficiency, with the resulting friction losses being major contributor to the transmission efficiencies reported in Section~\ref{sec:SISO&SIMOoperation}.


The clutches act to couple to the input shaft when turned ON with 1000~V applied and begin to disengage when turned OFF with 0~V applied. An Analog Technologies AHVAC5KV1MBT was used to provide the high voltage signal to the clutches. To control clutch engagement, a SPDT pushbutton (NKK Switches
LP01) is used to switch the clutching electrode connection between ground and high voltage. For dynamic tests, a Tektronix AFG3252 signal generator in conjunction with opto-couplers (OPTO-150) configured in a half bridge is used to supply the high voltage signal.

The total weight of our multiplexing transmission with four outputs is 1.12~kg, not including the high voltage power supply, hardware used to fix the system to the optic table and the weight of the motor. The relevant system specifications are recorded in Table~\ref{tab:systemspecs3}, along with individual output unit specifications showing how the system would scale in terms of size and weight for each additional output.

\begin{table}
    \centering
    \caption{physical specifications}
    \begin{tabular}{|c|c|}
    \hline
        Multiplexer Dimensions & 12 x 254 x 17 cm\\
    \hline
        Multiplexer Weight & 1.12~kg\\
   \hline
        Single Unit Size & 17 x 10 x 28 cm\\
    \hline
        Single Unit Weight & 110 g\\
    \hline
    \end{tabular}
    \label{tab:systemspecs3}
\end{table}

\section{Mechanical Multiplexer Single Unit Performance}





For dynamic robotic manipulation, the response time and transient behavior of output transitions are critical metrics to assess the system capability and operating limits of the mechanical multiplexer. Quasi-static experiments are also useful for finding the maximum output capability. Here, both the static and dynamic performance of a single unit is evaluated. Single units are combined in parallel to make up the complete mechanical multiplexer with four outputs. 

The input power from the motor to the mechanical multiplexer is measured with a ATO-TQS-DYN-200 rotary torque sensor. For force acting on the output leadscrew nut, a Mxmoonfree HP-500 force gauge is rigidly fixed to the leadscrew output nut and the optics table. For dynamic tests, a Qilichuangan A1S3 load cell with the load cell amplifier C2A3 was used for the increased sampling rate of 1280~Hz. Engaging the rightward clutch will move to increase the strain on the force gauge. Engaging the leftward clutch will move to increase the compression on the force gauge.

\subsection{Maximum Output Capability}

An input motor speed of 2 revolutions per minute (rpm) was used to find the approximate maximum single output force. The maximum measured output force was 212 N, with clutch slip observed as the failure mode at a measured input motor torque of 1.84~N·m. The average power consumption of the active electrostatic clutch under load was 2.4~mW. 

To measure the maximum output speed, the leadscrew nut was actuated without an external load. A 120~ms rightward pulse followed by a 120~ms leftward pulse with the output tracked with a 240 frames per second video with distance references. A 5~ms delay was inserted between the clutch commands to allow the residual clutch holding force to dissipate and prevent interference with the opposing clutch. Over 10 trials with an input shaft rpm of 360, the average output speed over the 120~ms ON pulse, for both left and rightward translation was 69.5~mm/s ± 2.4~mm/s.

\subsection{Dynamic Response Characterization}
\label{sec:DynamicPerformance}


During initial dynamic testing, we found that the compliance of the 3D printed brackets dominated the performance. To isolate the performance of the mechanical multiplexing components (leadscrew and clutches), a single unit was constructed using aluminum bracketing and ball bearings that support the thrust and rotation of the leadscrew, following the recommendations in \cite{gobilda3501Series}.

The increased frame stiffness reduced structural compliance and enabled measurement of the mechanical multiplexing architecture dynamic performance rather than that of the mounting structure. We provide the same performance metrics with our 3D printed implementation to compare to.

\begin{figure*}[t]
\centering
\includegraphics[width=0.99\linewidth]{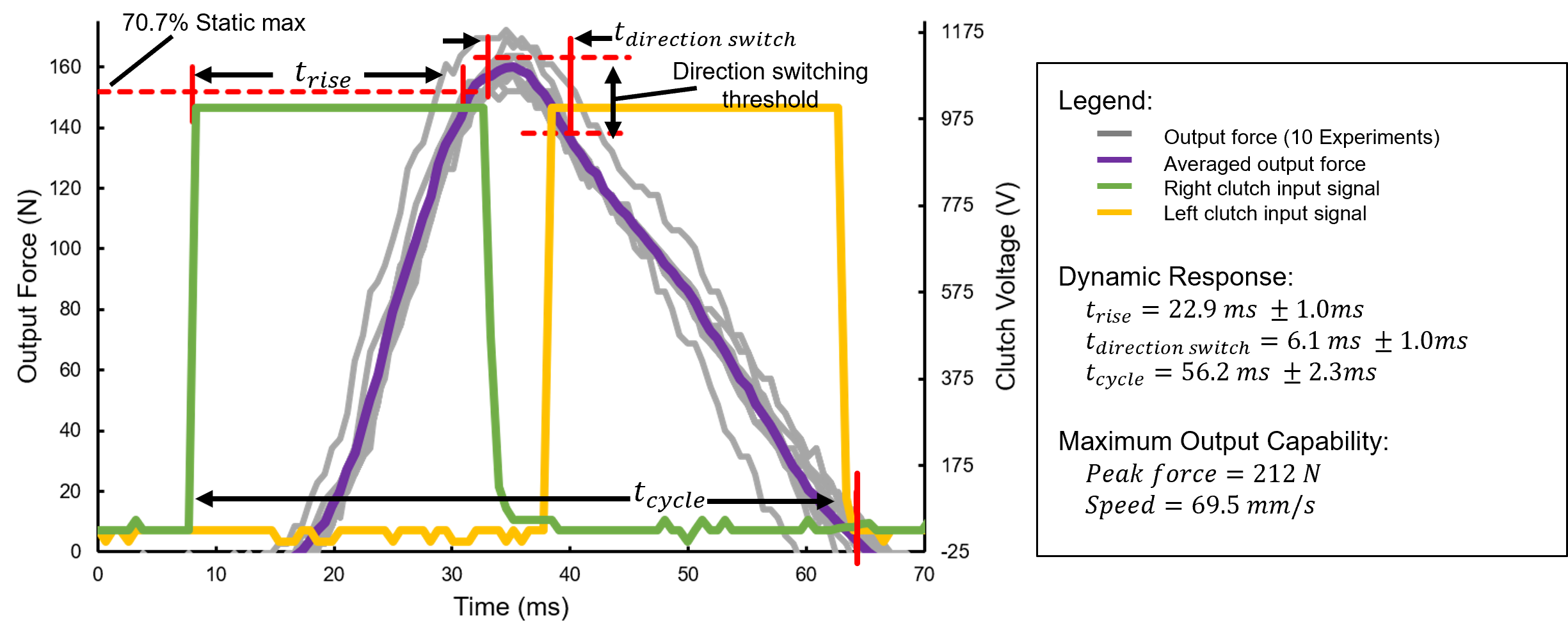}
\vspace{-5mm}
\caption{\label{fig:Dynamics}Dynamic response characterization of response time and transient behavior of a single unit with stiffened support structure (described in Section~\ref{sec:DynamicPerformance}). The evaluated metrics are rise time $t_{rise}$, direction-switching delay $t_{ds}$ and total cycle time $t_{cycle}$. "Direction switching threshold" denotes $85\%$ of the previous maximum during rightward translation, ensuring the output has switched directions. Performance metrics were determined from the response to a pair of consecutive input pulses: a 25~ms rightward clutch ON pulse followed by a leftward 25~ms clutch ON pulse after a 5~ms delay.}
\end{figure*}

Dynamic performance metrics were determined from the response to a pair of consecutive input pulses (shown in Fig.~\ref{fig:Dynamics}): a 25~ms rightward clutch ON pulse followed by a leftward 25~ms clutch ON pulse after a 5~ms delay, so the residual right clutch holding torque does not prevent the leftward clutch from properly engaging. The input rpm was set to 120. The metrics evaluated were rise time $t_{rise}$, load direction-switching delay $t_{ds}$ and total cycle time $t_{cycle}$. Consistent with bandwidth and transient-response characterization conventions, a force threshold corresponding to the 3 dB point (70.7~\% of the maximum output force) was used to define loaded dynamic metrics. Metrics were computed from $n = 10$ trials. $t_{rise}$ is the time required to increase force from 0~N to 152~N, $t_{cycle}$ is the time between the initial right clutch pulse and the return to 0~N, and $t_{ds}$ is the delay between peak force output and the force decreasing below 85~\% of the preceding this peak force, ensuring the leadscrew output nut has changed translation direction to the left. Definitions, average values, and standard deviations for dynamic response metrics are shown in Fig.~\ref{fig:Dynamics}.

Compared to the unit with a 3D printed support structure (also averaged over $n = 10$ trials), response times decreased by approximately an order of magnitude. $t_{rise}$ decreased from 274~ms to 22.9~ms and $t_{cycle}$ from 590~ms to 56.2~ms. In contrast, $t_{ds}$ decreased only modestly from 11.3~ms to 6.1~ms. Unlike $_{trise}$ and $t_{cycle}$, which are strongly influenced by structural compliance, the transmission is already tensioned during direction switching. As a result, the measured delay is dominated by clutch disengagement and engagement dynamics rather than frame deformation. These results indicate that the intrinsic dynamic response of the mechanical multiplexing architecture is substantially faster than suggested by the initial 3D printed prototype, with the majority of the observed delay attributable to structural compliance rather than clutch or transmission dynamics.

\section{Complete System: SISO \& SIMO Multiplexing}
\label{sec:SISO&SIMOoperation}

\begin{figure}[t]
\centering
\includegraphics[width=0.99\linewidth]{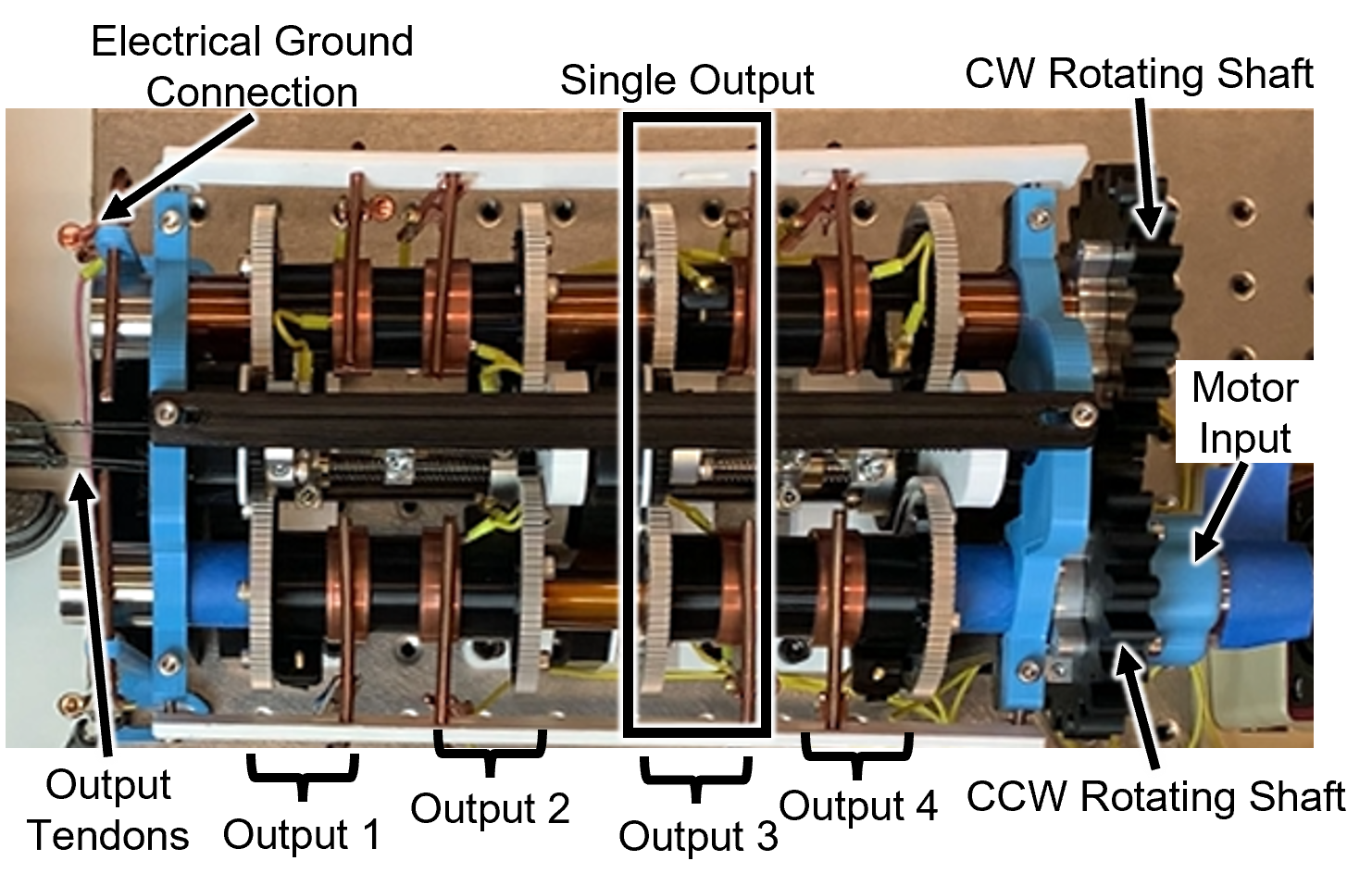}
\vspace{-8mm}
\caption{\label{fig:multiplexer} The full multiplexer is comprised of four single units, depicted in Fig.~\ref{fig:single-unit}, tiled in parallel along the same input shafts to independently and simultaneously control four outputs, including fully-actuated DoF control. }
\end{figure}

Here, we demonstrate SISO and SIMO operation of our multiplexer with four outputs built with four single units (Fig.~\ref{fig:single-unit}) tiled in parallel along shared input shafts, shown in Fig.~\ref{fig:multiplexer}. Full actuation of output DoF is demonstrated in Section~\ref{sec:DynamicPerformance} and Section~\ref{sec:Drill}.

An ATO-TQS-DYN-200 rotary torque sensor mounted between the motor and mechanical multiplexer measures the input mechanical power from the motor to the mechanical multiplexer.  To evaluate our system in this section, 2.27~kg weights are attached to each output leadscrew nut using a 1.33~mm diameter Dyneema cord, draped over a 25.4~mm stainless steel shaft to hang vertically while maintaining the tension of the cord parallel along the multiplexer.


\subsection{SISO Operation}

\begin{figure}[t]
\centering
\includegraphics[width=0.99\linewidth]{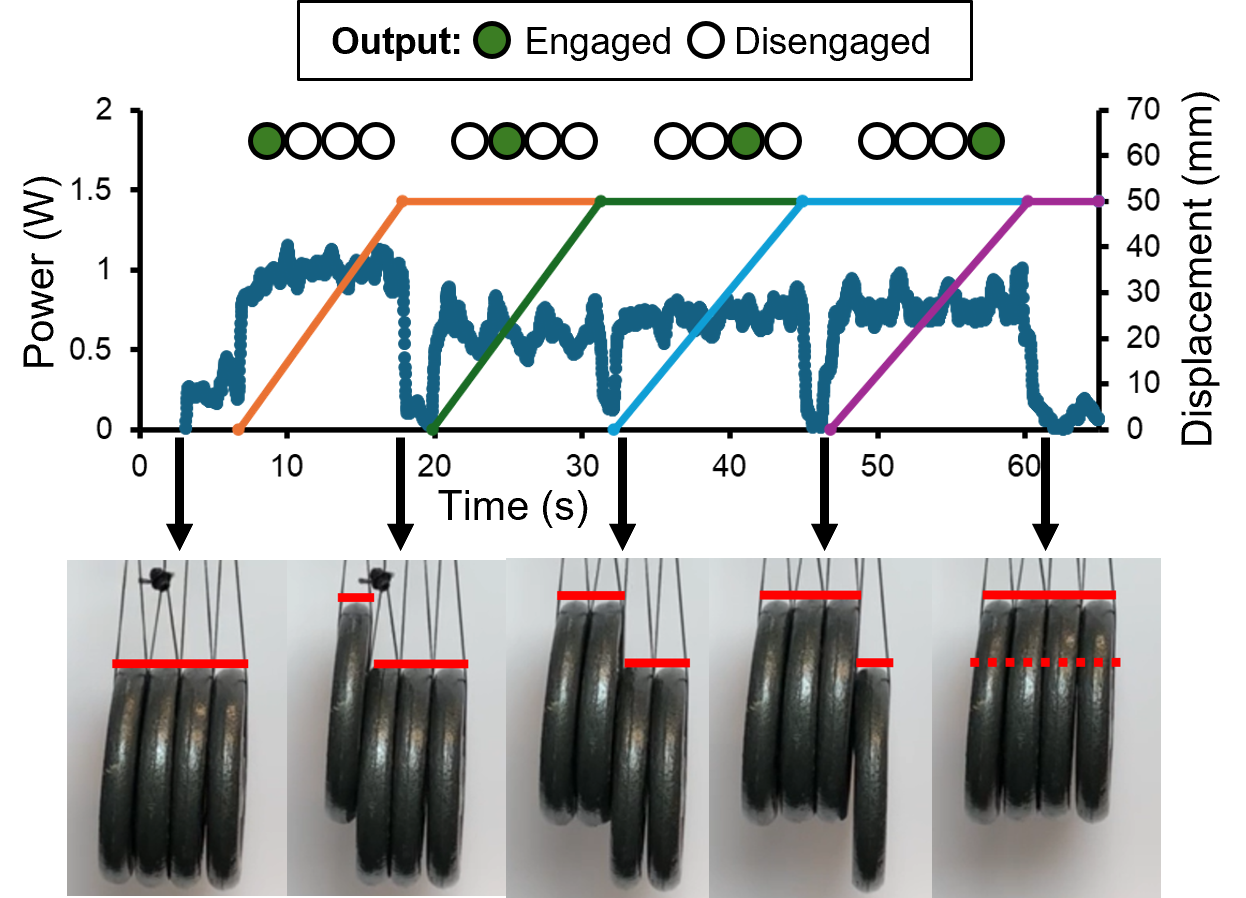}
\vspace{-6mm}
\caption{\label{fig:SISO} In SISO multiplexing, a single motor is clutched to different outputs, each individually in time. The top circles indicate which output is selected.  As clutches are individually activated, the corresponding 2.27 kg weight is lifted. The average output power for lifting the weight between all outputs was 0.77~W. Transmission efficiencies are recorded in Table~\ref{tab:SISO-Eff}.}
\vspace{-.2cm}
\end{figure}

To demonstrate SISO multiplexing, only one output is active at a time, as shown in a time-lapse in Fig.~\ref{fig:SISO}. Each 2.27~kg weight is vertically moved 50~mm with an input shaft speed of 18~rpm. Fig.~\ref{fig:SISO} shows how the transmitted mechanical power is switched between the different outputs over time. The time division of SISO allows for control of four individual outputs, with an average output power of 0.77~W. To control four outputs, a typical implementation would require four motors, each capable of 0.77~W. Recorded in Table~\ref{tab:SISO-Eff}, the energy efficiency of the mechanical multiplexer excludes motor efficiency by integrating the mechanical power curve from the rotary torque sensor summed with the electrical power consumption of the clutch and comparing to the total system output energy of moving the weight against gravity.

The relatively low efficiencies reported in Tables~\ref{tab:SISO-Eff} and \ref{tab:SIMO-Efficientcies} primarily result from prototype implementation choices, including compliant 3D-printed support structures and the absence of dedicated bearing surfaces, further discussed and characterized in Section~\ref{sec:DynamicPerformance}. The active electrostatic clutch consumed only 2.4~mW under load. Therefore, the reported efficiencies should not be interpreted as a fundamental limitation of electrostatic clutch-based mechanical multiplexing.

\begin{table}
    \centering
    \caption{Efficiencies calculated from power graph in Fig.~\ref{fig:SISO}}
    \begin{tabular}{|c|c|c|c|c|}
    \hline
       Output Actuated  & 1 & 2 & 3 & 4 \\
    \hline
        Transmission Efficiency & 10.3\% & 15.9\% & 12.2\% & 10.7\%\\
    \hline 
    \end{tabular}
    \label{tab:SISO-Eff}
\end{table}

\subsection{SIMO Operation}

\begin{figure}[t]
\centering
\includegraphics[width=0.99\linewidth]{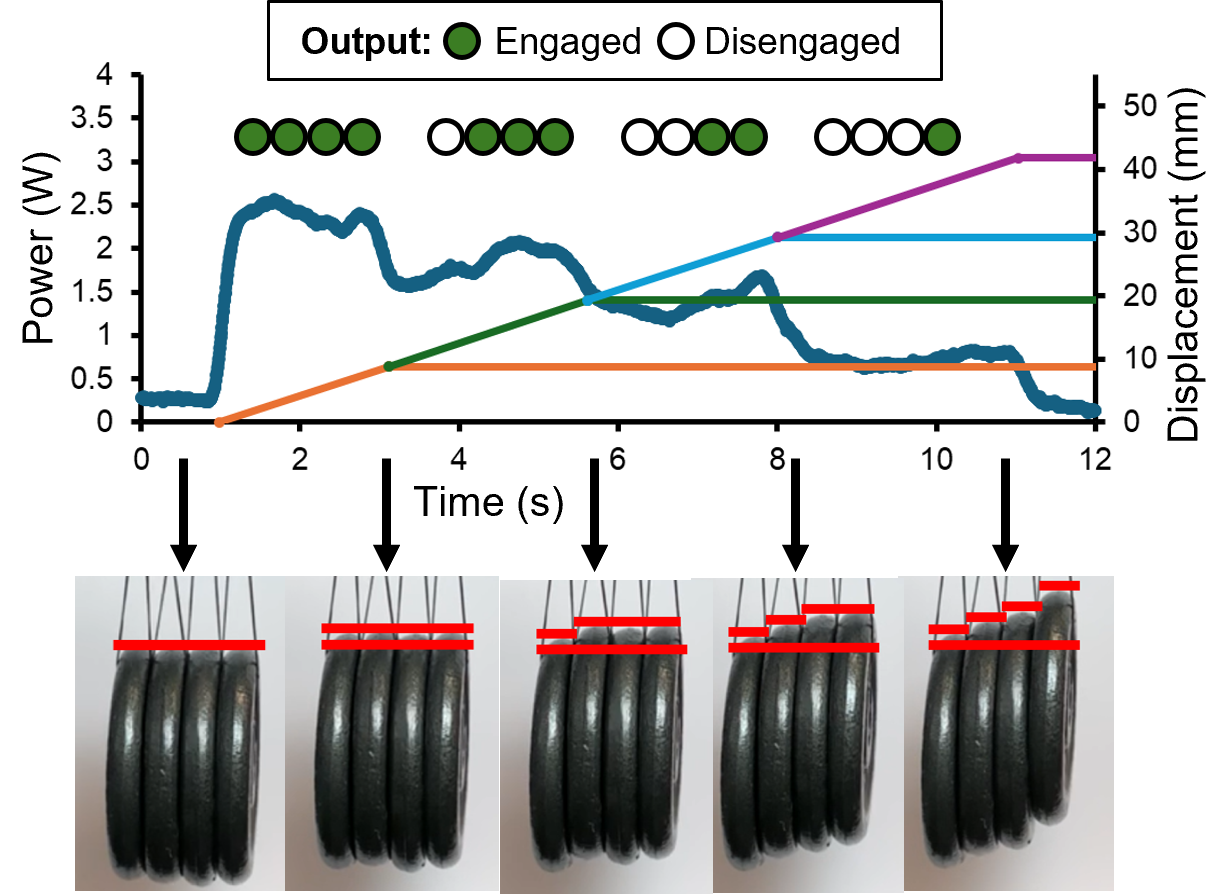}
\vspace{-6mm}
\caption{\label{fig:SIMO} In SIMO operation, a single motor controls four different outputs (connected to 2.27~kg weights) both independently and simultaneously without changing the system configuration shown in Fig.~\ref{fig:multiplexer}. The top circles indicate which outputs are engaged. As the number of outputs engaged decreases, so does the input power to the system. Efficiencies for each group of engaged outputs is shown in Table~\ref{tab:SIMO-Efficientcies}.}
\vspace{-.2cm}
\end{figure}

To demonstrate SIMO, a single motor independently actuates multiple outputs simultaneously. As in the SISO demonstration in Fig.~\ref{fig:SISO}, 2.27~kg weights are moved vertically, this time with a different grouping of outputs engaged to show independence. This SIMO control is shown in the time-lapse Fig.~\ref{fig:SIMO}. Initially, all four clutches on the CCW rotating shaft are ON and coupled to the shaft and moved together. Then only three, two, and then one outputs are selected over time. For each phase observed in Fig.~\ref{fig:SIMO}, the transmitted motor power stair steps down as individual outputs are held in place, decreasing the total system load on the motor. The energy efficiencies for each grouping of actuated outputs are recorded in Table ~\ref{tab:SIMO-Efficientcies} and calculated using the same method as for SISO. The efficiency is highest with all outputs engaged as the clutch drag torque is reduced, consistent with \ref{eq:Fi_ratio}.

\begin{table}
    \centering
    \caption{Efficiencies calculated from power graph in Fig.~\ref{fig:SIMO}}
    \begin{tabular}{|c|c|c|c|c|}
    \hline
       Outputs Actuated  & 1, 2, 3, 4 & 2, 3, 4 &  3, 4 & 4 \\
    \hline
        Transmission Efficiency & 16.4\% & 15.3\% & 13.3\% & 12.0\%\\
    \hline 
    \end{tabular}
    \label{tab:SIMO-Efficientcies}
\end{table}

\section{Increasing Force Capacity with SISO}
\label{sec:ForceMultiplicationWithSISO}

SISO multiplexing increases the individual output force by allocating the full system power to a single output rather than distributing it across multiple outputs. Because the output leadscrews are nonbackdrivable, force can be maintained at previously actuated outputs, allowing system-wide force to accumulate over time. For a four-output system with symmetric performance, the ideal SISO force increase is by a factor of 4 compared to simultaneous actuation (SIMO), as predicted by (\ref{eq:SISO/SIMO}). Relative to a conventional multi-actuator architecture employing four actuators with equivalent total torque capacity, SISO operation also provides a fourfold increase in the maximum force available to an individual output.

To validate this behavior, a motor stall torque of 0.74~N·m was used as a system-wide load limit while output tension was measured using a Mxmoonfree HP-500 force gauge. SIMO operation served as the baseline and was also representative of a conventional multi-actuator architecture with four actuators and equivalent total torque capacity. During SIMO testing, all four outputs were engaged until motor stall occurred. Outputs were combined and equalized at motor stall through a whippletree mechanism \cite{Smith-1988-Whippletree}. For SISO testing, each output was engaged individually.

\begin{figure}[t]
\centering
\includegraphics[width=0.99\linewidth]{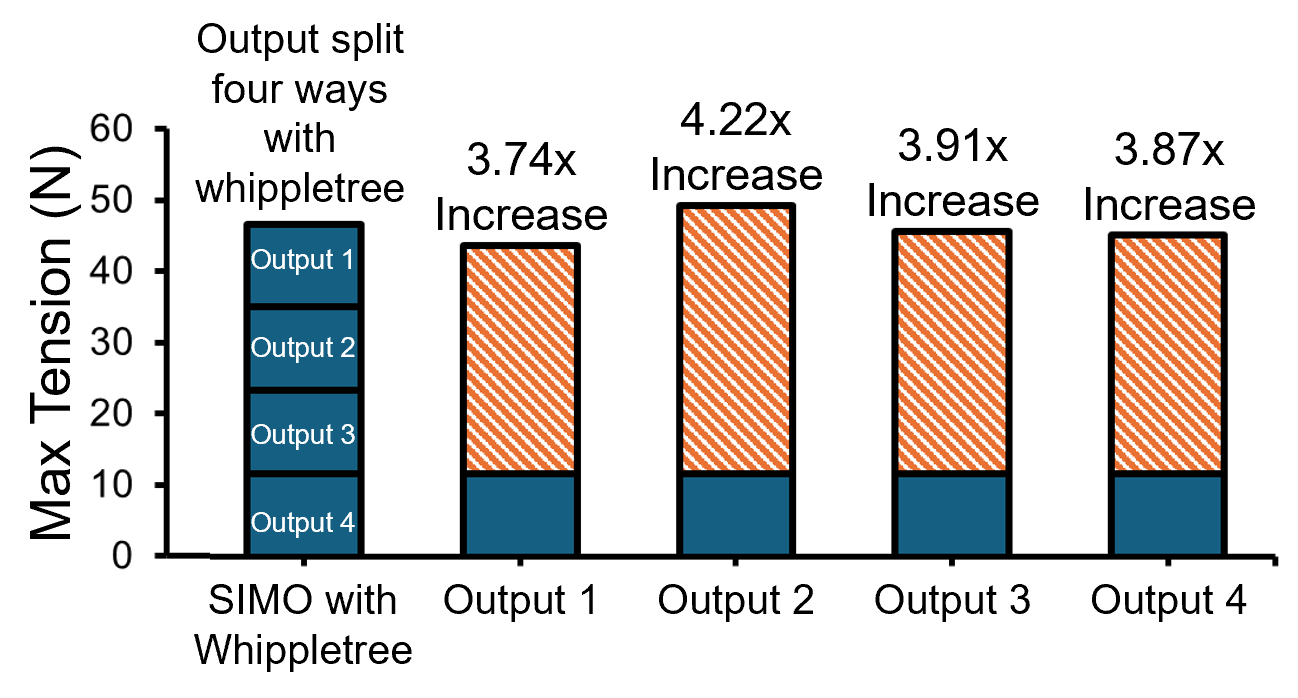}
\vspace{-8mm}
\caption{Data summary of experiment described in Section~\ref{sec:ForceMultiplicationWithSISO}, comparing the maximum individual output force during SIMO (blue boxes) and SISO (orange striped boxed) operation under the same system-wide torque limit. SISO increased maximum individual output force by a factor of 3.94, closely matching the theoretical prediction of 4 from \eqref{eq:SISO/SIMO}.}
\label{fig:WhippleTree}
\end{figure}

For SIMO operation, the average maximum combined output tension was 46.76~N ± 1.34~N (n = 10), with each individual output contributing an equal fourth of 11.69~N. For SISO operation, the average maximum individual output tension across all four outputs was 46.01~N ± 2.05~N (n = 10 per output). The resulting individual output force increased by a factor of 3.94, closely matching the ideal value of 4 and the N-fold scaling prediction of (\ref{eq:SISO/SIMO}). These results validate that SISO multiplexing can allocate the full system power to an individual output. The results are summarized in Fig.~\ref{fig:WhippleTree}. 

\section{Multiplexing Commercial Robotic Hand}

Here we demonstrate the advantages of our electrostatic clutch-based mechanical multiplexer with the commercial tendon-driven Seed Robotics RH8D hand provided without motors and with tendons exposed for connection to our system. The leadscrew output nuts were connected with fishing line (Power Pro, 100 lb) to tendons that manipulate the index, middle, and thumb, with ring and pinky fingers tied together.

\subsection{SIMO Positioning and SISO To Increasing Grip Strength}
\label{sec:SIMO-Positioning-SISO-Increase}

\begin{figure*}[t]
\centering
\includegraphics[width=0.99\linewidth]{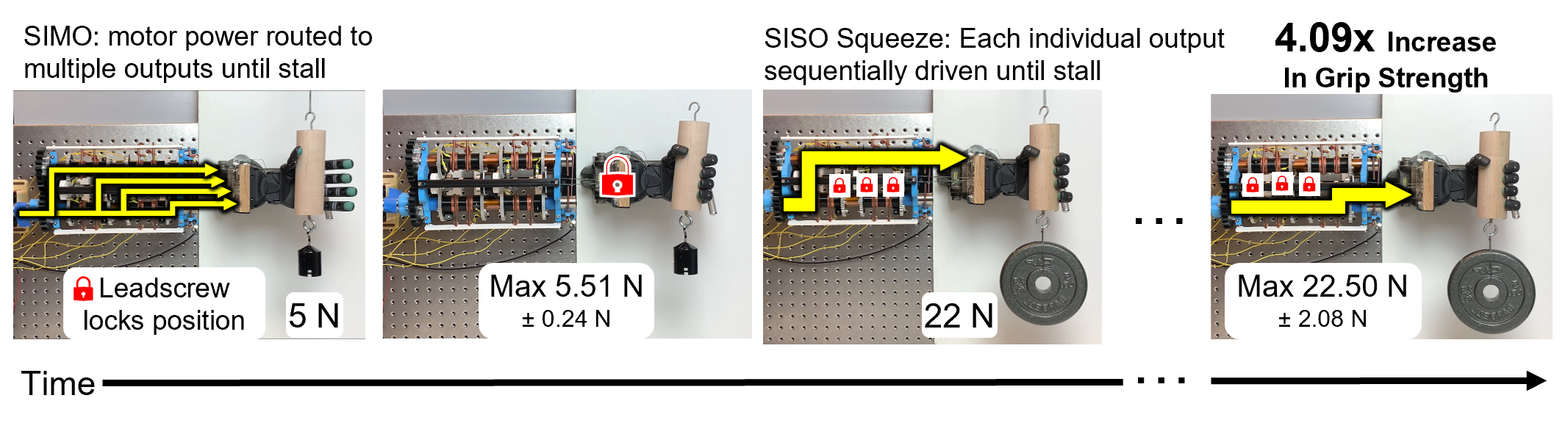}
\vspace{-6mm}
\caption{\label{fig:Hand-Squeeze} Squeezing procedure to increase the grip strength of a commercial robot hand. SIMO multiplexing is used to move fingers into position. With SISO multiplexing full motor power is routed sequentially to each finger maximizing force output. The non-backdrivability of leadscrews maintains the pressure of the non-active fingers. Using this squeezing sequence, the grip strength increased by a factor of 4.09 from 5.51~N to 22.50~N (n = 10 trials).}
\end{figure*}

\begin{figure*}[t]
\centering
\includegraphics[width=0.99\linewidth]{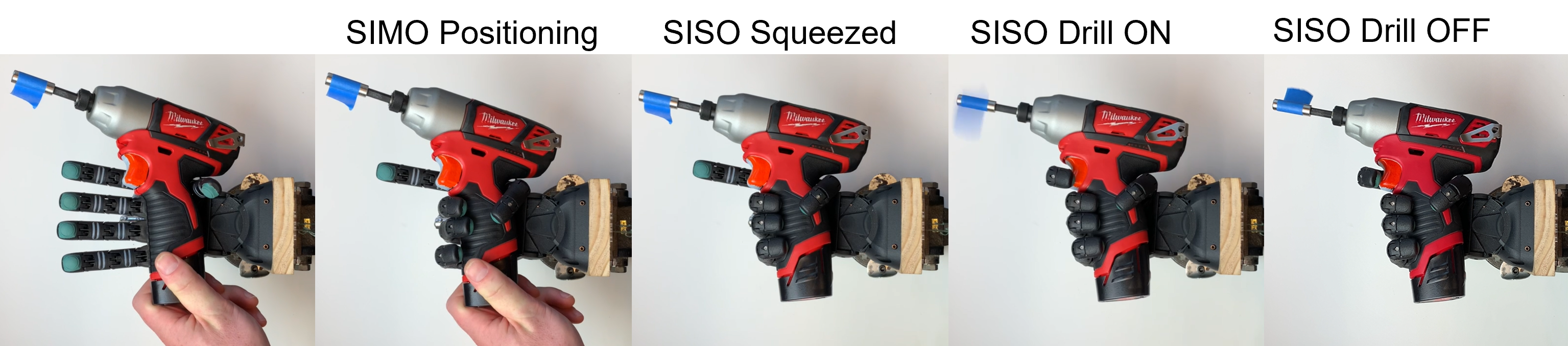}
\vspace{-2mm}
\caption{\label{fig:Drill} Demonstration of drill operation using only one motor. First, SIMO multiplexing is used to move fingers into position. Next, SISO is used iteratively to increase the grip strength on the drill as shown in Fig.~\ref{fig:Hand-Squeeze}. The non-backdrivability of the leadscrews maintains the grip on the drill. SISO multiplexing is then used with the pointer finger to turn the drill ON and OFF.}
\end{figure*}

SIMO multiplexing is used to position multiple fingers independently and simultaneously. Section~\ref{sec:ForceMultiplicationWithSISO} validates the prediction in \eqref{eq:SISO/SIMO} that using SISO, individual and system-wide output forces can increase by approximately the number of outputs, in our case 4, compared to SIMO with all outputs engaged (representative of a traditional robotic transmission  architecture). Here we show that this increased force capacity does not just work for our mechanical multiplexer in isolation but also for the driven robotic system.

To measure grip strength, a 51.4~mm diameter wooden cylinder with a metal hook screwed parallel into the end is placed in the robotic hand in a vertical position. Using SIMO, all fingers are driven simultaneously to grip the cylinder until the torque on the driveshaft causes the motor to stall at 0.74~N·m running at 18~rpm. Once stalled, the motor is turned off and clutches are disengaged with fingers held in position by the leadscrews. A digital force gauge (Mxmoonfree model HP-500) records the maximum force required to pull the cylinder out of the hand perpendicular to the finger orientation, as shown in Fig.~\ref{fig:Hand-Squeeze}. 

Demonstrating the increased force capacity benefit of SISO, after all fingers stall the motor with SIMO, the clutches are disengaged, allowing the motor to return to speed. The total motor power is now sequentially switched between each output over time until motor stall, increasing each finger force. The wooden cylinder remained undeformed, allowing the fingers to maintain force and position.

Using only SIMO with all outputs active, on average, 5.51 ± 0.24 N (n = 10 trials) force was required to remove the cylinder from the robot hand. With the addition of sequential squeezing with SISO, the grip strength increased by a factor of 4.09 to 22.50~N ± 2.08~N (n = 10 trials). The small increase over the theoretical prediction in \eqref{eq:SISO/SIMO} is attributed to the increased engagement of the robotic hand's rubber tipped fingers at higher forces. 

The same squeezing sequence was also used to increase the holding capacity in a horizontal farmer's carry position. The maximum payload we achieved was 11.34~kg (111.2~N), the highest recorded value compared to other five-fingered tendon-driven robotic hands, shown in Fig.~\ref{fig:Hero}.

\subsection{Operating Drill with Benefit from SIMO \& SISO}
\label{sec:Drill}

To show the real-world applicability of SIMO and SISO multiplexing, our novel transmission is used with the Seed Robotics hand to operate a drill (Milwaukee 12~V impact driver).  The Seed Robotics hand actuates each finger using a single tendon, causing the distal phalange to rotate at the Proximal Interphalangeal joint and pinch the trigger between the distal phalange and proximal phalange rather than pull with the middle phalange \cite{SunNing-2021-HandAnatomy}. To compensate for this kinematic shortcoming, a lighter pull trigger (NKK Switches LP01) was installed and painted orange in Fig.~\ref{fig:Drill}, decreasing the required actuation force from 16.9~N to 1.32~N. Drill operation is shown in Fig.~\ref{fig:Drill}. First, the middle, ring, pinky and thumb were moved into position using SIMO for the initial grip. SISO then increased each of these finger forces to further increase grip strength. Finally, SISO was used with the pointer finger to actuate the drill trigger ON and OFF. This procedure increased the drill vertical grip strength from
55.1~N to 118.76~N.


\section{Robotic Hand Comparison}
\label{sec:RoboticHandComparison}


Mechanical multiplexers used to operate robotic hands are found in \cite{Patrick-2023-InHandManipulation,Aukes-2014-hand,Wei-2023-SoftRobotic-Multiplexing,Zuo-2020-Mplex-SurgicalRobot,Kim-2022-Switchable-Cable,Xu-2024-MuxHand,Mathieu-2013-MechanicallyProgrammedMultiplexHand,Colgate-2026-Hand}. With \cite{Wei-2023-SoftRobotic-Multiplexing,Xu-2024-MuxHand,Aukes-2014-hand,Kim-2022-Switchable-Cable,Colgate-2026-Hand}, all of them notably use electrostatic clutches, provide performance metrics, and are directly compared to 22 five-finger tendon-driven robotic hands from academia and industry\cite{ShadowHand,MCRHandII,TendonDrivenDexterousHand,MiniXHand,Pisa-IIT-SoftHand2,Prensilia-IH2-Azzurra,qb-SoftHand-Industry,Robonaut2,X-Hand,TRX-Hand5,CEA-Dexterous-Hand,RoboRay,UNIPI-Hand,SmartHand,Keio-Hand,Vanderbuilt-Hand,X-Limb,SeedRobotics-RH8D,RUKA-Hand,ORCA-Hand,ETHZurich-Faive,DexHand,Xu-2024-MuxHand}. The robotic hands are compared in terms of holding capacity, hand cycle frequency (how many times a hand can fully close and open per second), size, weight, DoA and number of actuators. Power consumption is not compared due to differences in how it is reported. All comparison data collected are provided in the Appendices together with the reported power consumption divided into different reporting categories. Discussion of the role of non-backdrivability in robotic hands can be found in \cite{Belter-2013-NonbackdrivbabilityRobotHands,Montagnani-2015-NonbackdrivabilityinRobotHands,Belter-2013-AnthropomorphicRoboticHandReview}.


 
\begin{figure}[t]
\centering
\includegraphics[width=0.99\linewidth]{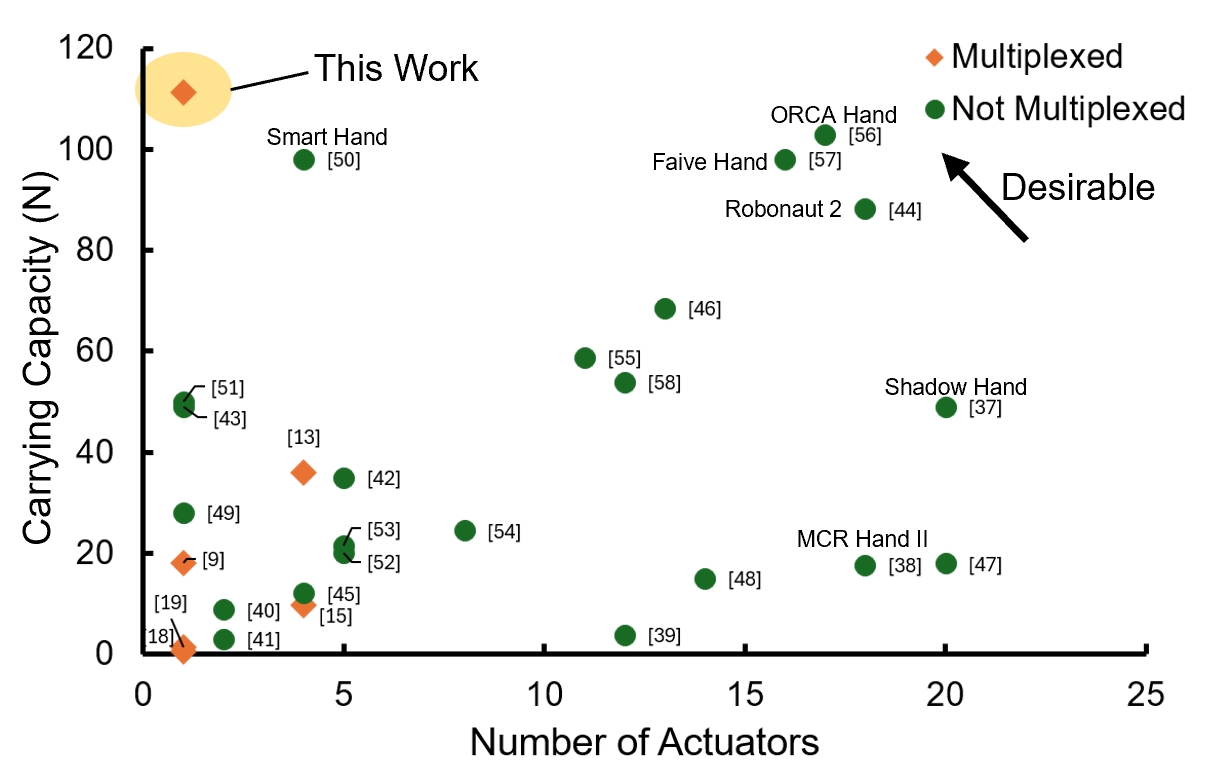}
\vspace{-8mm}
\caption{\label{fig:Holding-vs-Actuators} Comparison of the carrying capacity vs. number of actuators used of mechanically multiplexed and five-fingered tendon-driven robotic hands. Our work is highlighted demonstrating the highest recorded carrying capacity of 111.2~N in the category, using a single motor.}
\vspace{-4mm}
\end{figure}

Fig.~\ref{fig:Holding-vs-Actuators} compares robotic hand horizontal carrying capacity against the number of actuators used. As of publication our mechanical multiplexer allowed us to construct a robotic hand with the highest carrying capacity of five-fingered tendon-driven robotic hands while maintaining a considerably low maximum motor power consumption of 7.5~W, including electrostatic clutch power consumption of 9.6~mW. 

\begin{figure}[t]
\centering
\includegraphics[width=0.99\linewidth]{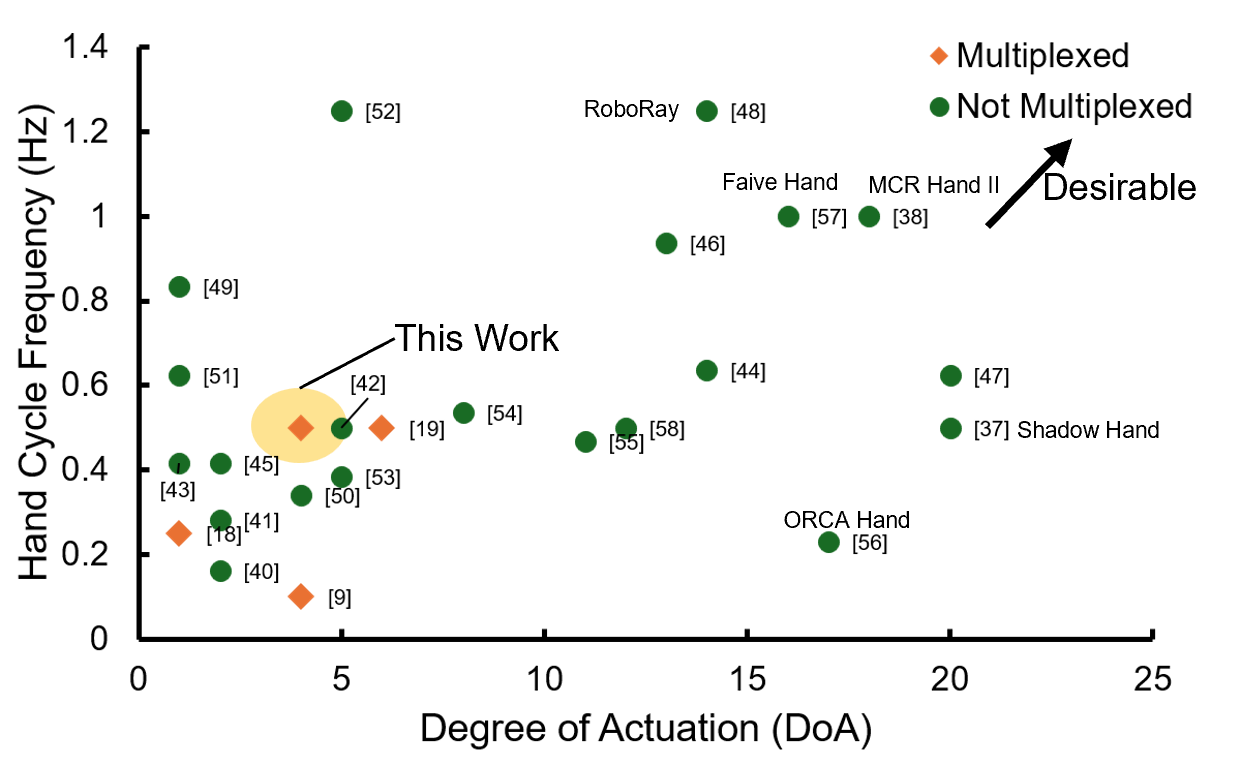}
\vspace{-8mm}
\caption{\label{fig:Speed-vs-DoA} Comparison of the hand cycle frequency (how many times a hand can fully close and open a second) vs. DoA. Values are either taken directly from the original sources or estimated from the maximum observed cycle frequency in publicly available hand operation videos. For highly articulated and dynamic robotic platforms, a fast speed with many DoA is desirable.}
\vspace{-4mm}
\end{figure}

\begin{figure}[t]
\centering
\includegraphics[width=0.99\linewidth]{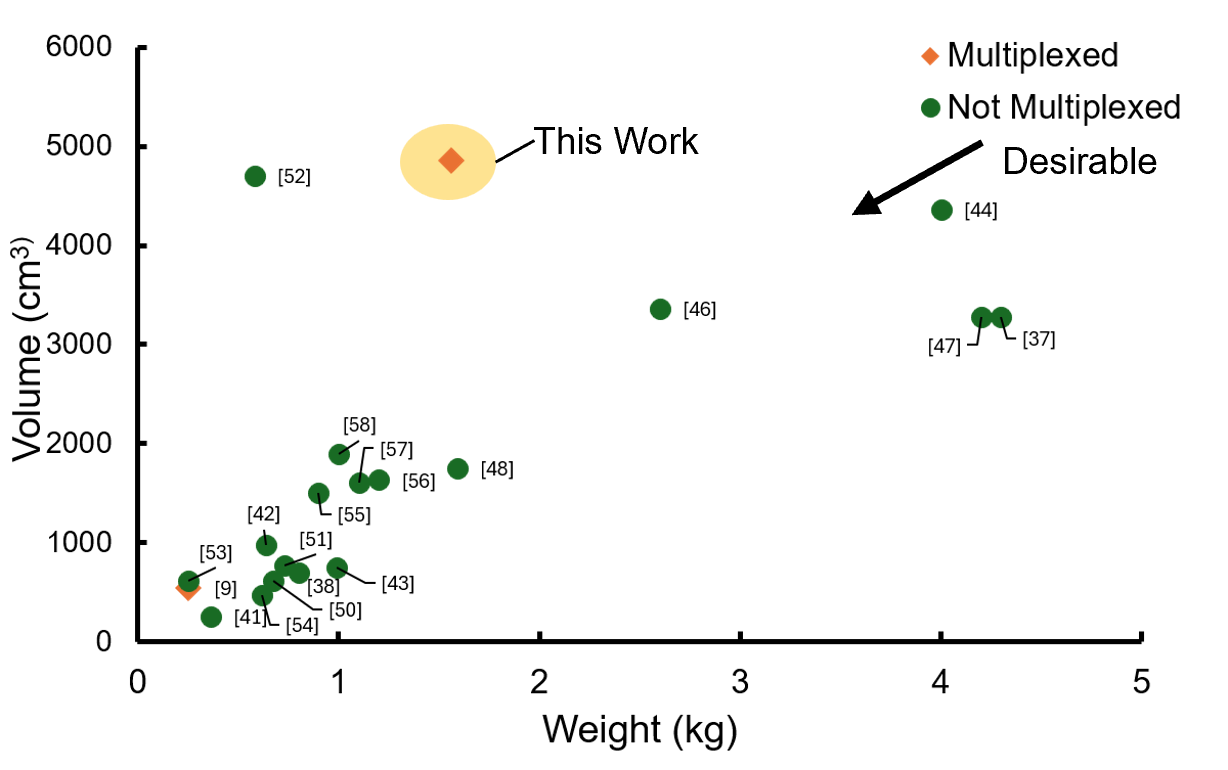}
\vspace{-8mm}
\caption{\label{fig:Volume-vs-Weight} Comparison of volume vs. weight of mechanically multiplexed and five-fingered tendon-driven robotic hands from both industry and academia. Our work is highlighted and indicates a significant improvement would be miniaturization.}
\vspace{-4mm}
\end{figure}

Figs.~\ref{fig:Speed-vs-DoA} and \ref{fig:Volume-vs-Weight} indicate where the most significant improvements could be made to our system. Fig.~\ref{fig:Speed-vs-DoA} compares the hand cycle frequency against DoA. Fig.~\ref{fig:Speed-vs-DoA} is a measure of how many different DoF can be actuated and at what frequency. For a highly dynamic system, it is desirable to have many DoA that can be actuated quickly. As indicated by Fig.~\ref{fig:Speed-vs-DoA}, a future improvement will be increasing the level of multiplexing to gain a larger DoA. Fig.~\ref{fig:Volume-vs-Weight} compares size and weigh. For comparison, motor weight and volume are added to the values reported in Table~\ref{tab:systemspecs3}.

\section{Conclusions}






This work presented an electrostatic capstan clutch-based mechanical multiplexer capable of SISO and SIMO operation while preserving fully-actuated control of each DoF. A single motor independently and simultaneously controlled four outputs without mechanical reconfiguration.

Experimental validation demonstrated output forces up to 212~N, speeds of up to 69.5~mm/s, and 152~N dynamic response times as low as 22.9~ms. A system-level model accurately predicted force-allocation behavior for the operating regimes of SISO and SIMO multiplexing.

Integration with a commercial robotic hand demonstrated a factor of 4.09 increase in grip strength and a horizontal carrying capacity of 111.2~N, the highest reported among five-fingered tendon-driven robotic hands.


Mechanical multiplexing can provide an alternative actuation paradigm for highly articulated robotic systems. These results demonstrate that mechanical multiplexing can reduce actuator count while preserving fully-actuated control of DoFs and enabling dynamic allocation of force capacity. Future work will focus on efficiency, miniaturization, increased multiplexing density, and task-dependent control strategies that transition between SISO and SIMO operation.

\vspace{-4mm}
\section*{Acknowledgments}
The authors thank Dr. Alex Langrock for assisting in the identification and acquisition of PBI for this work.
\vspace{-2mm}

\bibliographystyle{IEEEtran}
\bibliography{main}

\begin{IEEEbiography}{Timothy E. Amish}
Biography text here.
\end{IEEEbiography}

\begin{IEEEbiography}{Jeffrey T. Auletta}
Biography text here.
\end{IEEEbiography}

\begin{IEEEbiography}{Chad C. Kessens}
Biography text here.
\end{IEEEbiography}

\begin{IEEEbiography}{Joshua R. Smith}
Biography text here.
\end{IEEEbiography}

\begin{IEEEbiography}{Jeffrey I. Lipton}
Biography text here.
\end{IEEEbiography}

\end{document}